\def\ARXIVVERSION{1}

\PassOptionsToClass{nohead,nofoot}{asmeconf}

\documentclass[colorlinks,upint,subscriptcorrection,varvw,hyphenate,balance, main=english]{asmeconf}

\ifdefined\ARXIVVERSION
    \hypersetup{
        pdfauthor={Fadel M. Megahed and Ying-Ju Chen},
        pdftitle={Beyond CLIP: Calibrated (Few-Shot) Defect Detection With System One Decision Models},
        pdfkeywords={Clef-Flash, CLIP, computer vision,
            few-shot classification, industrial image inspection,
            Jev, System One models},
        pdfsubject={Research preprint}
    }
\else
    \hypersetup{
        pdfauthor={Fadel M. Megahed and Ying-Ju Chen},
        pdftitle={Beyond CLIP: Calibrated (Few-Shot) Defect Detection With System One Decision Models},
        pdfkeywords={Clef-Flash, CLIP, computer vision,
            few-shot classification, industrial image inspection,
            Jev, System One models},
        pdfsubject={MSEC 2027-204306 DRAFT Paper}
    }
\fi

\usepackage[export]{adjustbox}
\usepackage{algorithm}
\usepackage{algpseudocode}
\usepackage{amsthm}
\usepackage{array}
\usepackage{enumitem}
\usepackage{fontawesome5}
\usepackage[figuresright]{rotating}
\usepackage{soul}
\usepackage{stfloats}
\usepackage{tikz}
\usepackage{eso-pic}

\usetikzlibrary{shapes.geometric, arrows, arrows.meta, positioning, backgrounds}
\newcolumntype{M}{>{\centering\arraybackslash}m{0.65cm}}

\allowdisplaybreaks

\usepackage{threeparttable,makecell,tikz,tabularx,xcolor}
\usepackage[normalem]{ulem}
\definecolor{jev}{HTML}{00829B}
\newcommand{\latx}[1]{\fpeval{(ln(#1)/ln(10) - 0.69897)*0.38}in}
\newcommand{\latglyph}[5]{\begin{tikzpicture}[baseline=-0.6ex]
  \foreach \t in {10,100,1000} \draw[black!15, line width=0.3pt] (\latx{\t},-3.6pt) -- (\latx{\t},3.6pt);
  \draw[black!55, line width=0.4pt] (\latx{#1},0) -- (\latx{#2},0) (\latx{#4},0) -- (\latx{#5},0);
  \fill[black!25] (\latx{#2},-2.3pt) rectangle (\latx{#4},2.3pt);
  \draw[black, line width=0.9pt] (\latx{#3},-3pt) -- (\latx{#3},3pt);
  \path (0,0) (\latx{2000},0);
\end{tikzpicture}}
\newcommand{\lataxis}{\begin{tikzpicture}[baseline=0pt, font=\fontsize{6}{7}\selectfont]
  \draw[black!50, line width=0.3pt] (0,0) -- (\latx{2000},0);
  \foreach \t in {10,100,1000} {\draw[black!50, line width=0.3pt] (\latx{\t},0) -- (\latx{\t},-2pt);
    \node[anchor=north, inner sep=1pt] at (\latx{\t},-2pt) {\t};}
\end{tikzpicture}}

\begin{document}

\ifdefined\ARXIVVERSION
\else
    \AddToShipoutPictureFG*{%
      \begin{tikzpicture}[remember picture,overlay]
        \node[anchor=north,font=\bfseries\Large]
          at ([yshift=-0.25in]current page.north) {DRAFT};
      \end{tikzpicture}%
    }

    \ConfName{\vspace*{0.2in}Proceedings of the ASME 2027\linebreak
    22nd International Manufacturing Science and Engineering Conference}
    \ConfAcronym{MSEC2027}
    \ConfDate{June 21--June 25, 2027}
    \ConfCity{Corvallis, Oregon}
    \PaperNo{MSEC2027-204306}
\fi

\title{Beyond CLIP: Calibrated (Few-Shot) Defect Detection With System One Decision Models}

\SetAuthors{%
	Fadel M. Megahed\affil{1}\CorrespondingAuthor{fmegahed@miamioh.edu},
	Ying-Ju Chen\affil{2}
	}

\SetAffiliation{1}{Farmer School of Business, Miami University, Oxford, OH 4050, USA}
\SetAffiliation{2}{College of Arts and Sciences, University of Dayton, Dayton, OH 45469, USA}

\maketitle

\versionfootnote{Draft of \today.}


\begin{abstract}
Cameras and vision sensors are widely used for product and process inspection in manufacturing environments. Traditionally, this process required thousands of images of both nominal and defective products to train such models. Recent research shows that vision transformer (ViT) models, such as CLIP, can reduce this requirement to 50–100 images per nominal and defective class. However, collecting even 50 defective images can be difficult in high-yield or low-volume manufacturing processes. In this paper, we examine the utility of two recently proposed structured decision models (\texttt{Jev} and \texttt{Clef-Flash}) in applications using as few as $k=2$ reference images per class at inference. In the process, we provide three main contributions: (a) a framework for $k-$shot learning of these models for image classification/inspection applications; (b) an agentic approach for automatically refining the text descriptions that guide image classification in these decision models; and (c) an approach to calibrate the resulting defect probabilities to improve their reliability for inspection decisions. We test these contributions in five manufacturing case studies. Using two reference images per class, both \texttt{Jev} and \texttt{Clef-Flash} outperform CLIP's test accuracies with the same references (0.75--1.00 [\texttt{Jev}] and 0.71--0.99 [\texttt{Clef-Flash}] vs 0.67--0.84 [CLIP]). Furthermore, we show that our agentic refinement of text descriptions significantly improves accuracy across 6 of 8 model-application combinations, and our calibration approach reduces the Brier score in 13 of the 15 model-application configurations. To support the reproducibility of our work, we provide our replication code and have created the \faRProject \ ``decisions'' package for evaluating additional models. We also provide a publicly available web app that allows practitioners to inspect their own images with no code.   
\end{abstract}

\section{Introduction}
Computer vision has become an integral component of quality assurance in modern manufacturing environments \citep{colosimo2024statistical}, enabling 100\% inspection of parts at high production speeds \citep{megahed2011review}. The standard approach is to train a custom model for each inspection task, which requires task-specific images and possibly annotations \citep{jeong2023winclip}. However, in \textit{Industry 4.0/5.0} environments, obtaining sufficiently large training image sets has become increasingly difficult since: (a) smart manufacturing systems tend to produce very few defective parts \citep{psarommatis2026zero}, which limits the availability of images of defective parts; and (b) the shift from mass production to mass customization and personalization \citep{hu2013evolving} means that many product variants undergo too few production cycles to generate sufficiently large training datasets.

Large pretrained vision transformers offer a potential solution to this data bottleneck. Vision-language models, such as CLIP, pair a vision transformer image encoder with a text encoder and are pretrained on 400 million image-text pairs \citep{radford2021learning}. This enables zero-shot classification from text descriptions alone, without task-specific training images \citep{jeong2023winclip}. However, CLIP often fails as a zero-shot classifier in manufacturing applications since representative defect images are rare in its web-scale training data \citep{megahed2026clip}. 

\begin{figure}[htb!]
    \centering
    \includegraphics[width=0.99\linewidth, frame]{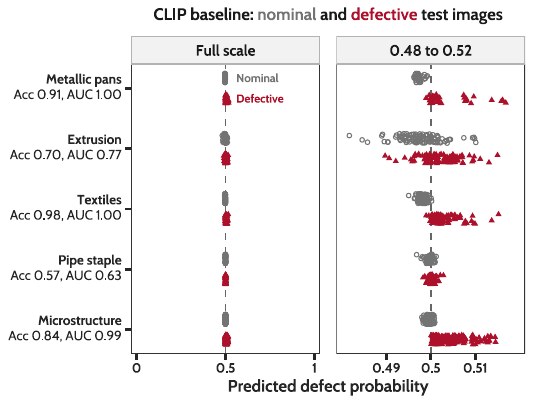}

    \vspace{-0.5\baselineskip}
    
    \caption{Compression of CLIP-predicted defect probabilities across the manufacturing case studies of \cite{megahed2026clip}.}
    \label{fig:compression}

    \vspace{-\baselineskip}
\end{figure}

Using a few labeled reference images, Megahed \textit{et al.} \cite{megahed2026clip} improved CLIP's classification accuracy across several manufacturing case studies. For metallic pan surfaces, accuracy increased from 0.615 to 0.910 using 10 reference images per class. For stochastic textured surfaces (textiles), accuracy increased from 0.500 to 0.975 using 50 reference images per class. Smaller improvements were observed for 3D printing extrusion profiles (0.620 to 0.700) and automotive pipe staple assemblies (0.500 to 0.580), both using 50 reference images per class. Despite these gains, defect probabilities obtained by applying softmax to CLIP's similarity scores clustered near 0.5 for both nominal and defective images (FIGURE~\ref{fig:compression}). This clustering suggests substantial uncertainty despite the observed high classification accuracies, raising questions about their reliability for decision-making.

Recently proposed structured decision models (also known as \textit{System One} models, such as \texttt{Jev} \citep{typesafe2026introducing} and \texttt{Clef-Flash} \citep{cloudfare2025clef}) may address these limitations. These models can use input images and user-provided text descriptions to answer predefined questions by assigning probabilities to user-specified answer choices.  We hypothesize that these properties can make them appealing for manufacturing inspection applications, where users can define their inspection criteria through text descriptions and specify a set of possible outcomes (e.g., nominal or defective). However, it is unclear whether these properties translate well to manufacturing applications that require limited reference images. It is also important to evaluate the reliability of their probability estimates under these conditions, where such estimates can inform decisions to accept, reject, or further inspect a product.

In this paper, we investigate the utility of two structured decision models, \texttt{Jev} and \texttt{Clef-Flash}, for image-based quality inspection using as few as $k=2$ reference images per class. Specifically, we make three contributions:
\begin{enumerate}[label=(\arabic*), nosep]
    \item We develop a framework for $k$-shot learning with structured decision models for image classification and inspection applications. Our approach uses reference images at inference without requiring application-specific model training or fine-tuning.

    \item We introduce an agentic AI approach to automatically refine the text descriptions that guide image classification. This approach aims to improve classification accuracy while reducing the need to manually develop descriptions for different inspection tasks.

    \item We propose an approach to calibrate the obtained classification probabilities so they better reflect the uncertainty in the model's predictions.
\end{enumerate}

We evaluate these contributions using five manufacturing case studies previously investigated in \cite{megahed2026clip}. These cases cover metallic surfaces, 3D printing extrusion profiles, stochastic textured surfaces, automotive assemblies, and microstructure images. We compare \texttt{Jev} and \texttt{Clef-Flash} with CLIP in two main settings: (a) matched labeled-image budgets and (b) matched reference-image configurations. We also examine whether agentically refined text descriptions improve classification accuracy and whether probability calibration reduces the Brier score.

\section{(Few-Shot) Inspection Framework for System One Decision Models}

\begin{figure*}[bth!]
  \centering

  \definecolor{methodink}{HTML}{263640}
  \definecolor{methodgray}{HTML}{4A5359}
  \definecolor{methodline}{HTML}{CDD4D8}
  \definecolor{methodpurple}{HTML}{7154A1}
  \definecolor{methodteal}{HTML}{187D80}
  \definecolor{methodorange}{HTML}{C2603B}
  \definecolor{methodblue}{HTML}{517F9B}

  \newcommand{\methodstack}[3]{%
    \begin{scope}[shift={(#1,#2)}]
      \draw[draw=#3!45,fill=white,line width=.6pt] (.13,.13) rectangle (.93,.73);
      \draw[draw=#3!60,fill=white,line width=.6pt] (.06,.06) rectangle (.86,.66);
      \draw[draw=#3,fill=white,line width=.7pt] (0,0) rectangle (.8,.6);
      \fill[#3!16] (.07,.07) rectangle (.73,.53);
      \draw[#3!65,line width=.6pt] (.12,.13) -- (.30,.33) -- (.43,.22) -- (.60,.40) -- (.69,.28);
      \fill[#3!70] (.23,.43) circle (.045);
    \end{scope}%
  }

  \newcommand{\methodmodel}[3]{%
    \begin{scope}[shift={(#1,#2)}]
      \draw[draw=#3,line width=.9pt,fill=white,rounded corners=1.5pt] (0,0) rectangle (1.5,1.5);
      \foreach \s in {.25,.5,.75,1,1.25}{
        \draw[#3!65,line width=.6pt]
          (-.13,\s)--(0,\s) (1.5,\s)--(1.63,\s)
          (\s,-.13)--(\s,0) (\s,1.5)--(\s,1.63);
      }
      \foreach \a in {.38,.75,1.12}{
        \foreach \b in {.38,.75,1.12}{
          \draw[#3!23,line width=.45pt] (.32,\a)--(.75,\b) (.75,\a)--(1.18,\b);
        }
      }
      \foreach \x in {.32,.75,1.18}{
        \foreach \y in {.38,.75,1.12}{
          \filldraw[fill=white,draw=#3,line width=.65pt] (\x,\y) circle (.075);
        }
      }
    \end{scope}%
  }
  \fbox{%
    \resizebox{\dimexpr\textwidth-2\fboxsep-2\fboxrule\relax}{!}{%
    \begin{tikzpicture}[
        x=1cm, y=1cm,
        font=\sffamily\fontsize{9}{11}\selectfont,
        text=methodink,
        flow/.style={-{Stealth[length=2.0mm,width=1.3mm]},draw=methodgray,line width=.9pt},
        fine/.style={draw=methodline,line width=.6pt},
        label/.style={align=center,inner sep=0pt},
        small/.style={font=\sffamily\fontsize{8}{9.5}\selectfont},
        letter/.style={anchor=west,inner sep=0pt,font=\sffamily\bfseries\fontsize{13}{14}\selectfont},
        heading/.style={anchor=west,inner sep=0pt,font=\sffamily\bfseries\fontsize{10}{12}\selectfont}
      ]
      \path[use as bounding box] (0,.2) rectangle (18.8,-14.15);

      \node[letter]  at (0,0)     {a};
      \node[heading] at (.55,0)   {Define the task};
      \node[letter]  at (4.8,0)   {b};
      \node[heading] at (5.35,0)  {Refine the descriptions};
      \node[letter]  at (13.05,0) {c};
      \node[heading] at (13.6,0)  {Calibrate and validate};
      \draw[fine] (4.5,-.55)--(4.5,-9.15);
      \draw[fine] (12.65,-.55)--(12.65,-9.15);

      \node[label,small,text=methodgray] at (2.05,-.65) {An extrusion example};
      \node[label,text=methodteal]   at (1.03,-1.2) {Nominal};
      \node[label,text=methodorange] at (3.07,-1.2) {Defective};
      \foreach \x/\y/\file/\assetcolor in {
          1.03/-2.35/extrusion-nominal-1/methodteal,
          1.03/-4.35/extrusion-nominal-2/methodteal,
          3.07/-2.35/extrusion-defective-1/methodorange,
          3.07/-4.35/extrusion-defective-2/methodorange}{
        \node[inner sep=0pt] at (\x,\y)
          {\includegraphics[width=1.72cm,height=1.72cm]{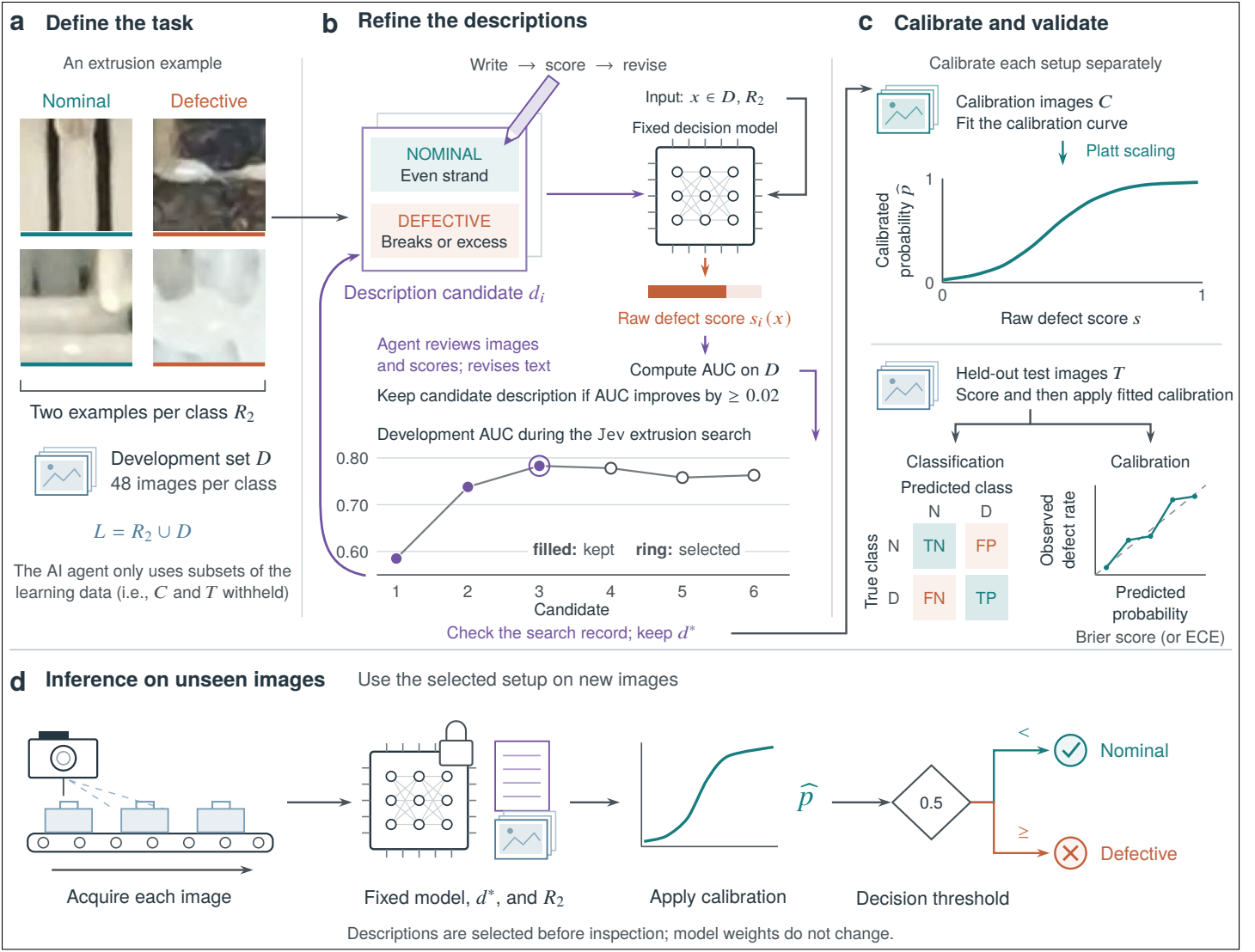}};
        \draw[\assetcolor,line width=1.8pt] (\x-.86,\y-.91)--(\x+.86,\y-.91);
      }
      \draw[methodgray,line width=.65pt] (.17,-5.5)--(.17,-5.67)--(3.93,-5.67)--(3.93,-5.5);
      \node[label] at (2.05,-6.02) {Two examples per class $R_2$};
      \methodstack{.4}{-7.27}{methodblue}
      \node[anchor=west,inner sep=0pt,align=left] at (1.55,-6.92)
        {Development set $D$\\\textcolor{methodgray}{48 images per class}};
      \node[label,text=methodblue] at (2.05,-7.85) {$L=R_2\cup D$};
      \node[label,small,text=methodgray] at (2.20,-8.62)
        {The AI agent only uses subsets of the\\learning data (i.e., $C$ and $T$ withheld)};
      \draw[flow] (4.03,-3.0)--(5.22,-3.0);

      \node[label,small,text=methodgray] at (8.6,-.65)
        {Write \ $\rightarrow$ \ score \ $\rightarrow$ \ revise};

      \draw[methodline,fill=white,line width=.65pt] (5.66,-1.4) rectangle (8.18,-3.6);
      \draw[methodpurple!75,fill=white,line width=.85pt] (5.43,-1.62) rectangle (7.95,-3.82);
      \fill[methodteal!10] (5.55,-1.77) rectangle (7.83,-2.6);
      \node[label,small,text=methodteal] at (6.69,-2.025) {NOMINAL};
      \node[label,small] at (6.69,-2.345) {Even strand};
      \fill[methodorange!9] (5.55,-2.79) rectangle (7.83,-3.62);
      \node[label,small,text=methodorange] at (6.69,-3.045) {DEFECTIVE};
      \node[label,small] at (6.69,-3.365) {Breaks or excess};

      \begin{scope}[shift={(7.75,-1.7)},rotate=-37]
        \fill[methodpurple!18] (-.10,0) rectangle (.10,1.03);
        \draw[methodpurple,line width=.75pt]
          (-.10,0)--(-.10,1.03)--(.10,1.03)--(.10,0)--(0,-.21)--cycle;
        \fill[methodpurple] (0,-.21)--(-.05,-.10)--(.05,-.10)--cycle;
      \end{scope}
      \node[label,text=methodpurple] at (6.7,-4.18) {Description candidate $d_i$};
      \draw[flow,draw=methodpurple] (8.26,-2.64)--(9.77,-2.64);

      \methodmodel{9.95}{-3.42}{methodink}
      \node[label,font=\sffamily\fontsize{7.5}{9}\selectfont] at (10.7,-1.62)
        {Fixed decision model};
      \node[label,font=\sffamily\fontsize{7.5}{9}\selectfont] at (10.7,-1.17)
        {Input: $x\in D$, $R_2$};
      \draw[flow] (11.95,-1.17)--(12.25,-1.17)--(12.25,-2.65)--(11.65,-2.65);
      \draw[flow,draw=methodorange] (10.7,-3.62)--(10.7,-3.92);
      \fill[methodorange!15] (9.83,-4.24) rectangle (11.57,-4.05);
      \fill[methodorange]    (9.83,-4.24) rectangle (11.02,-4.05);
      \node[label,small,text=methodorange] at (10.7,-4.56) {Raw defect score $s_i(x)$};
      \draw[flow,draw=methodpurple] (10.7,-4.82)--(10.7,-5.12);
      \node[label,small,text=methodink] at (10.7,-5.36) {Compute AUC on $D$};
      \draw[flow,draw=methodpurple] (12.15,-5.36)--(12.4,-5.36)--(12.4,-6.45);

      \draw[flow,draw=methodpurple,line width=1.05pt]
        (5.48,-8.51) .. controls (4.85,-8.4) and (4.78,-7.9) .. (4.78,-7.3)
        -- (4.78,-4.35) .. controls (4.78,-3.98) and (5.03,-3.67) .. (5.4,-3.57);
      \node[anchor=west,align=left,inner sep=0pt,text=methodpurple,small] at (5.65,-5.12)
        {Agent reviews images\\and scores; revises text};
      \node[anchor=west,align=left,inner sep=0pt,small] at (5.65,-5.76)
        {Keep candidate description if AUC improves by $\geq0.02$};

      \node[anchor=west,inner sep=0pt,small,text=methodink] at (5.65,-6.35)
        {Development AUC during the \texttt{Jev} extrusion search};
      \draw[methodgray,line width=.6pt] (5.65,-8.5)--(12.0,-8.5);
      \foreach \v/\y in {0.60/-8.14,0.70/-7.42,0.80/-6.7}{
        \draw[methodline,line width=.45pt] (5.65,\y)--(12.0,\y);
        \node[anchor=east,inner sep=2pt,small,text=methodgray] at (5.6,\y) {\v};
      }
      \draw[methodgray!75,line width=.85pt]
        (5.95,-8.248)--(7.05,-7.1464)--(8.15,-6.8224)
        --(9.25,-6.8584)--(10.35,-7.0024)--(11.45,-6.9664);
      \foreach \x/\y in {5.95/-8.248,7.05/-7.1464,8.15/-6.8224}{
        \filldraw[fill=methodpurple,draw=white,line width=.5pt] (\x,\y) circle (.09);
      }
      \foreach \x/\y in {9.25/-6.8584,10.35/-7.0024,11.45/-6.9664}{
        \filldraw[fill=white,draw=methodgray,line width=.8pt] (\x,\y) circle (.085);
      }
      \draw[methodpurple,line width=.8pt] (8.15,-6.8224) circle (.155);
      \foreach \x/\i in {5.95/1,7.05/2,8.15/3,9.25/4,10.35/5,11.45/6}{
        \node[label,small,text=methodgray] at (\x,-8.75) {\i};
      }
      \node[label,small,text=methodink] at (8.65,-9.03) {Candidate};
      \node[label,small,text=methodgray,fill=white,inner sep=2pt] at (9.65,-8.13)
        {\textbf{filled:} kept \quad \textbf{ring:} selected};
      \node[label,small,text=methodpurple] at (8.65,-9.4)
        {Check the search record; keep $d^*$};
      \draw[flow] (11.1,-9.4)--(12.87,-9.4)--(12.87,-1.02)--(13.28,-1.02);

      \node[label,small,text=methodgray] at (15.94,-.65) {Calibrate each setup separately};

      \methodstack{13.35}{-1.7}{methodteal}
      \node[anchor=west,inner sep=0pt,small,align=left] at (14.55,-1.38)
        {Calibration images $C$\\Fit the calibration curve};
      \draw[flow,draw=methodteal] (16.2,-1.82)--(16.2,-2.2);
      \draw[methodgray,line width=.7pt] (14.35,-4.0)--(18.35,-4.0) (14.35,-4.0)--(14.35,-2.4);
      \node[anchor=east,inner sep=2pt,small,text=methodgray]  at (14.3,-4.0)   {0};
      \node[anchor=east,inner sep=2pt,small,text=methodgray]  at (14.3,-2.4)   {1};
      \node[anchor=north,inner sep=2pt,small,text=methodgray] at (14.35,-4.02) {0};
      \node[anchor=north,inner sep=2pt,small,text=methodgray] at (18.35,-4.02) {1};
      \draw[methodteal,line width=1.4pt]
        plot[smooth] coordinates {
          (14.36,-3.96) (14.82,-3.89) (15.28,-3.74) (15.72,-3.45)
          (16.18,-3.05) (16.62,-2.75) (17.08,-2.57) (17.55,-2.49) (18.28,-2.46)};
      \node[rotate=90,label,small,text=methodink] at (13.62,-3.2)
        {Calibrated\\probability $\widehat p$};
      \node[label,small,text=methodink] at (16.33,-4.55) {Raw defect score $s$};
      \node[anchor=west,inner sep=0pt,small,text=methodteal] at (16.55,-2.0) {Platt scaling};
      \draw[methodline,line width=.65pt] (13.25,-4.95)--(18.65,-4.95);

      \methodstack{13.35}{-5.95}{methodblue}
      \node[anchor=west,inner sep=0pt,small,align=left] at (14.55,-5.58)
        {Held-out test images $T$\\Score and then apply fitted calibration};
      \draw[flow] (15.7,-5.95)--(15.7,-6.18)--(14.55,-6.18)--(14.55,-6.48);
      \draw[flow] (15.7,-6.18)--(17.55,-6.18)--(17.55,-6.48);
      \node[label,small,text=methodink] at (14.55,-6.75) {Classification};
      \node[label,small,text=methodink] at (17.55,-6.75) {Calibration};

      \node[label,small,text=methodink]  at (14.57,-7.15)  {Predicted class};
      \node[label,small,text=methodgray] at (14.225,-7.52) {N};
      \node[label,small,text=methodgray] at (15.025,-7.52) {D};
      \node[label,small,text=methodgray] at (13.6,-8.05)   {N};
      \node[label,small,text=methodgray] at (13.6,-8.85)   {D};
      \node[label,small,text=methodink,rotate=90] at (13.25,-8.45) {True class};
      \fill[methodteal!16]  (13.9,-7.7) rectangle (14.55,-8.4);
      \fill[methodorange!9] (14.7,-7.7) rectangle (15.35,-8.4);
      \fill[methodorange!9] (13.9,-8.5) rectangle (14.55,-9.2);
      \fill[methodteal!16]  (14.7,-8.5) rectangle (15.35,-9.2);
      \node[label,small,text=methodteal]  at (14.225,-8.05) {TN};
      \node[label,small,text=methodorange] at (15.025,-8.05) {FP};
      \node[label,small,text=methodorange] at (14.225,-8.85) {FN};
      \node[label,small,text=methodteal]  at (15.025,-8.85) {TP};

      \draw[methodgray,line width=.6pt] (16.7,-8.5)--(18.4,-8.5) (16.7,-8.5)--(16.7,-7.12);
      \draw[methodgray!65,dashed,line width=.6pt] (16.7,-8.5)--(18.4,-7.12);
      \draw[methodteal,line width=.9pt]
        (16.87,-8.388)--(17.21,-7.965)--(17.55,-7.905)--(17.89,-7.344)--(18.23,-7.292);
      \foreach \x/\y in {16.87/-8.388,17.21/-7.965,17.55/-7.905,17.89/-7.344,18.23/-7.292}{
        \fill[methodteal] (\x,\y) circle (.04);
      }
      \node[label,small,text=methodink,rotate=90] at (16.12,-7.81) {Observed\\defect rate};
      \node[label,small,text=methodink]  at (17.55,-8.95) {Predicted\\probability};
      \node[label,small,text=methodgray] at (17.55,-9.48) {Brier score (or ECE)};

      \draw[methodline,line width=.8pt] (0,-9.65)--(18.8,-9.65);
      \node[letter]  at (0,-10.13)   {d};
      \node[heading] at (.55,-10.13) {Inference on unseen images};
      \node[anchor=west,inner sep=0pt,text=methodgray] at (5.35,-10.13)
        {Use the selected setup on new images};

      \draw[methodink,line width=.85pt,rounded corners=1pt,fill=white] (.3,-11.03) rectangle (1.35,-11.62);
      \fill[methodink] (.45,-10.91) rectangle (.91,-11.03);
      \draw[methodink,line width=.85pt] (.83,-11.32) circle (.19);
      \draw[methodink!45,line width=.65pt] (.83,-11.32) circle (.10);
      \draw[methodgray,line width=.7pt] (.83,-11.63)--(.83,-11.9);
      \draw[methodblue!60,dashed,line width=.55pt] (.92,-11.66)--(1.6,-12.1) (.92,-11.66)--(2.38,-12.1);

      \draw[methodink,line width=.8pt,rounded corners=3pt] (.28,-12.48) rectangle (4.05,-12.76);
      \foreach \x in {.52,1.08,1.64,2.2,2.76,3.32,3.83}{
        \draw[methodgray,line width=.6pt] (\x,-12.62) circle (.08);
      }
      \foreach \x in {.55,1.72,2.89}{
        \filldraw[fill=methodblue!13,draw=methodblue!75,line width=.7pt] (\x,-12.1) rectangle (\x+.72,-12.46);
        \draw[methodblue!75,line width=.8pt] (\x+.23,-12.1)--(\x+.23,-11.99)--(\x+.5,-11.99)--(\x+.5,-12.1);
      }
      \draw[flow] (.65,-13.04)--(3.75,-13.04);
      \node[label] at (2.15,-13.47) {Acquire each image};
      \draw[flow] (4.27,-12.0)--(5.27,-12.0);

      \methodmodel{5.55}{-12.72}{methodink}
      \draw[methodpurple,fill=white,line width=.7pt] (7.47,-11.06) rectangle (8.3,-12.13);
      \foreach \y in {-11.28,-11.49,-11.7,-11.91}{
        \draw[methodpurple!60,line width=.65pt] (7.59,\y)--(8.17,\y);
      }
      \methodstack{7.47}{-12.87}{methodblue}

      \draw[methodgray,line width=.8pt,rounded corners=1pt,fill=white] (6.62,-11.05) rectangle (7.12,-11.43);
      \draw[methodgray,line width=.8pt]
        (6.72,-11.05)--(6.72,-10.9) arc[start angle=180,end angle=0,radius=.15]--(7.02,-11.05);
      \node[label] at (7.0,-13.47) {Fixed model, $d^*$, and $R_2$};
      \draw[flow] (8.62,-12.0)--(9.39,-12.0);

      \draw[methodgray,line width=.6pt] (9.72,-12.68)--(11.82,-12.68) (9.72,-12.68)--(9.72,-11.08);
      \draw[methodteal,line width=1.25pt]
        plot[smooth] coordinates {
          (9.76,-12.6) (10.1,-12.52) (10.43,-12.24) (10.73,-11.65) (11.05,-11.29) (11.75,-11.15)};
      \node[label,text=methodteal,font=\sffamily\fontsize{13}{14}\selectfont] at (12.25,-11.92) {$\widehat p$};
      \node[label] at (10.9,-13.47) {Apply calibration};
      \draw[flow] (12.65,-12.0)--(13.56,-12.0);

      \draw[methodgray,line width=.8pt,fill=white]
        (14.18,-11.43)--(14.78,-12.0)--(14.18,-12.57)--(13.58,-12.0)--cycle;
      \node[label,small] at (14.18,-12.0) {0.5};
      \draw[flow,draw=methodteal]   (14.78,-12.0)--(15.14,-12.0)--(15.14,-11.2)--(16.0,-11.2);
      \draw[flow,draw=methodorange] (14.78,-12.0)--(15.14,-12.0)--(15.14,-12.78)--(16.0,-12.78);
      \node[label,small,text=methodteal]  at (15.6,-10.92) {$<$};
      \node[label,small,text=methodorange] at (15.6,-12.46) {$\geq$};

      \draw[methodteal,line width=.9pt,fill=methodteal!8] (16.32,-11.2) circle (.24);
      \draw[methodteal,line width=1.1pt] (16.2,-11.2)--(16.29,-11.29)--(16.46,-11.08);
      \draw[methodorange,line width=.9pt,fill=methodorange!7] (16.32,-12.78) circle (.24);
      \draw[methodorange,line width=1.1pt] (16.21,-12.67)--(16.43,-12.89) (16.21,-12.89)--(16.43,-12.67);
      \node[anchor=west,inner sep=0pt,text=methodteal]  at (16.77,-11.2)  {Nominal};
      \node[anchor=west,inner sep=0pt,text=methodorange] at (16.77,-12.78) {Defective};
      \node[label] at (14.2,-13.47) {Decision threshold};
      \node[label,small,text=methodgray] at (9.4,-14.04)
        {Descriptions are selected before inspection; model weights do not change.};
    \end{tikzpicture}%
  }%
  }%

  \vspace{-0.5\baselineskip}

  \caption{Overview of our $k-$shot learning framework for using decision models (e.g., \texttt{Jev} and \texttt{Clef-Flash}) to inspect manufactured parts.}
  \label{fig:cmp-framework}

  \vspace{-.5\baselineskip}
  
\end{figure*}

FIGURE~\ref{fig:cmp-framework} provides an overview of our proposed framework for leveraging \textit{System One}-like decision models for 100\% visual inspection of manufactured parts. The framework, which does not involve retraining those models, consists of four sequential stages: (a) defining the inspection task by providing a choice-based question to the model, a preliminary plain language description of the choices (i.e., nominal and defective classes), and a few sample images for each class; (b) refining (if needed) the provided plain language description of the choices using an agentic AI loop; (c) calibrating the obtained classification probabilities from the prior step to ensure that they can support reliable decision making; and (d) using the AI-improved plain language descriptions, the reference images, and the fitted calibration function to inspect new images (note that the descriptions, image references, and calibration function are fixed for testing/ deployment). 

\subsection{Defining the Visual Inspection Task}

Before defining our task, note that \textit{System One} models do not produce text outputs. Instead, their outputs are currently limited to three or four decision types, commonly called \textit{primitives}. For example, \textit{Jev} (as of \texttt{jev-1.13.0}) supports \citep{typesafe2026primitives}: (a) \textit{Noul}, which provides only a binary response to a question (i.e., 0 or 1); (b) \textit{Choice}, which is used to select from 2 or more predefined categories while providing the probabilities associated with each choice and a confidence score; and (c) \textit{Score}, for assigning a position based on a predefined, ordered categorical level. More recently, \texttt{Jev} introduced an experimental \textit{Box} primitive to identify object or region locations within an image \citep{typesafe2026box}. In any case, to use these \textit{System One} models, the user provides a question, the possible responses and their descriptions, and the input to be evaluated (e.g., an image). For visual inspection tasks, we assume these primitives support applications such as identifying defective products, classifying defect types, assessing defect severity, and locating defects within an image. In this paper, we focus on binary classification using the \textit{Choice} primitive. However, our framework can also support multiclass classification (e.g., nominal-, defect 1-, or defect 2-type classification).

The first stage of our framework involves defining the inspection task by providing: (a) a question asking the model to classify an image as nominal or defective; (b) preliminary plain-language descriptions of the two classes; and (c) a few reference images illustrating each class. Let $x$ denote an inspection image, $d=(d_{\mathrm{N}},d_{\mathrm{D}})$ the descriptions of nominal and defective appearances, and $R_k$ the set of (k) labeled reference images per class. For zero-shot applications, no reference images are provided (i.e., $R_0=\varnothing$). Given these inputs, the decision model (M) assigns a probability to each class. We use the probability assigned to the defective class as the raw defect score:

\vspace{-\baselineskip}

\begin{equation}
s(x;d,R_k)=M(x;d,R_k), \qquad 0\leq s(x;d,R_k)\leq 1.
\label{eq}
\end{equation}

FIGURE~\ref{fig:cmp-framework}(a) shows the first stage using the 3D printing extrusion application. In this example, we use two reference images per class $(R_2)$ and 48 development images per class $(D)$. We provide the reference images to the decision model alongside each inspection image, and we use the development images in the second stage to evaluate and improve the initial class descriptions. Together, these images constitute the learning data ($L$). Note that our approach does not retrain the decision model. Instead, we seek to improve its classification performance by refining the text descriptions while keeping the model's parameters fixed.

\subsection{Agentic Description Refinement}\label{sec:cmp-agent}

The second stage attempts to address the question: ``How can we (automatically) improve the semantic description of the \textit{System One} model (if needed) when we do not know its underlying decision-making mechanisms?'' We note that a description that is informative to manufacturing quality practitioners, or even to one \textit{System One} model, may be suboptimal for another model. The problem is that the decision-making process of either open-weight or closed-weight models is essentially a black box. To address this question, we use an agentic AI loop inspired by the iterative, evaluation-driven process of \texttt{autoresearch} \cite{karpathy2026autoresearch}. 

Our adaptation starts with a generic description of nominal and defective images. The agent iteratively refines these descriptions using the labeled development dataset $D$ and feedback from the \textit{System One} model. Specifically, for each candidate description pair $d_i$, the model estimates the probability of a defect, $s_i(x)$, for each image $x \in D$, using the same reference images $R_2$. An external scoring program calculates the area under the receiver operating characteristic curve (AUC) and summarizes the predicted defect probabilities by defect type. The AI agent then uses its built-in vision-language model to compare image–description pairs of misclassified and reference examples, revealing limitations in the current descriptions. To ensure meaningful gains and to reduce over-correction, we set a minimum AUC increase of 0.02 relative to the best-performing candidate. 
Algorithm~\ref{alg:description_refinement} summarizes the proposed procedure, while FIGURE~\ref{fig:cmp-framework}(b) illustrates its application to extrusion quality inspection. 

\begin{algorithm}[htb!]
\caption{Agentic refinement loop of semantic descriptions}
\label{alg:description_refinement}
\footnotesize
\begin{algorithmic}[1]
\Require Development data $D$, reference set $R_k$, decision model $M$, initial descriptions $d_1$,  maximum candidates $N_{\max}=6$, minimum improvement $\delta=0.02$
\Ensure Selected descriptions $d^*$, evaluation history $H$
\State Set $N_{\max}\gets 6$, $\delta\gets 0.02$
\State $d^*\gets d_1$, $A_{\mathrm{best}}\gets-\infty$,
       $H\gets\emptyset$

\For{$i=1,\ldots,N_{\max}$}
    \Statex \Comment{Evaluate the current descriptions}
    \ForAll{$x\in D$}
        \State $s_i(x)\gets M(x;d_i,R_k)$
    \EndFor
    \State $A_i\gets\operatorname{AUC}
           (\{(y_x,s_i(x)):x\in D\})$
    \State Summarize $s_i(x)$ by defect type
    \State Record $d_i$, $A_i$, scores, and rationale in $H$

    \Statex \Comment{Retain meaningful improvements}
    \If{$A_i\geq A_{\mathrm{best}}+\delta$}
        \State $d^*\gets d_i$
        \State $A_{\mathrm{best}}\gets A_i$
    \EndIf

    \If{$i=N_{\max}$}
        \State \textbf{break}
    \EndIf

    \Statex \Comment{Propose change(s) to the class descriptions}
    \State Inspect poorly ranked images, scores, and $R_k$
    \If{no promising revision is identified}
        \State \textbf{break}
    \EndIf
    \State Propose $d_{i+1}$ by making a targeted revision to the class descriptions.
    \State Document the rationale for the revision
\EndFor
\State \Return $d^*,H$
\end{algorithmic}

\end{algorithm}

\subsection{Probability Calibration}\label{sec:cmp-calibration}

The third stage of our framework addresses the need for reliable probability estimates in inspection decisions. As shown in FIGURE~\ref{fig:cmp-framework}(c), we apply Platt scaling \citep{platt1999probabilistic} to calibrate the raw defect scores using a separate set of labeled images ($C$). For each model–application–reference configuration, we estimate the calibrated defect probability as

\vspace{-.75\baselineskip}

\begin{equation}\label{eq:cmp-platt}
    \widehat p(x) = \sigma\left(a+b\,\operatorname{logit}(\widetilde s(x))\right),
\end{equation}

\vspace{-.25\baselineskip}

\noindent where $\sigma(z)=1/(1+e^{-z})$, $\widetilde s(x)$ is the raw defect score clipped to $[0.005,0.995]$ to avoid infinite logits, and $a$ and $b$ are parameters estimated from the calibration data. To improve numerical stability, we use smoothed calibration targets when estimating the Platt scaling parameters.

Note that the calibration images are independent of the data used to refine the class descriptions and are not used to select the reference images. We evaluate the fitted calibration function on the held-out test images ($T$), assessing classification performance with the confusion matrix and the reliability of the estimated defect probabilities with the Brier score. Finally, we use isotonic regression as a sensitivity analysis to examine whether our results depend on the choice of calibration method \citep{zadrozny2002transforming}.

\subsection{Inference on Unseen Images}\label{sec:cmp-inference}
The final stage of our framework involves using the selected model configuration to inspect previously unseen images, as illustrated in FIGURE~\ref{fig:cmp-framework}(d). For each new inspection image ($x$), we provide the decision model with the selected class descriptions ($d^*$) and the same reference images ($R_k$) used during development. The resulting raw defect score is then transformed into a calibrated probability, $\widehat p(x)$, using the fitted calibration function from the previous stage. We classify an image as defective if $\widehat p(x)\geq 0.5$ and nominal otherwise. Note that the decision model, class descriptions, reference images, and calibration function are fixed throughout this stage. Thus, our framework can inspect new images without retraining the underlying decision model or repeating the description refinement and calibration procedures.

\section{Experimental Setup}

\subsection{Five Manufacturing Case Studies and Data Allocation}
\label{sec:cmp-allocation}

\begin{figure}[htb!]
    \centering

    \vspace{-0.5\baselineskip}
    
    \includegraphics[width=.99\columnwidth, frame]{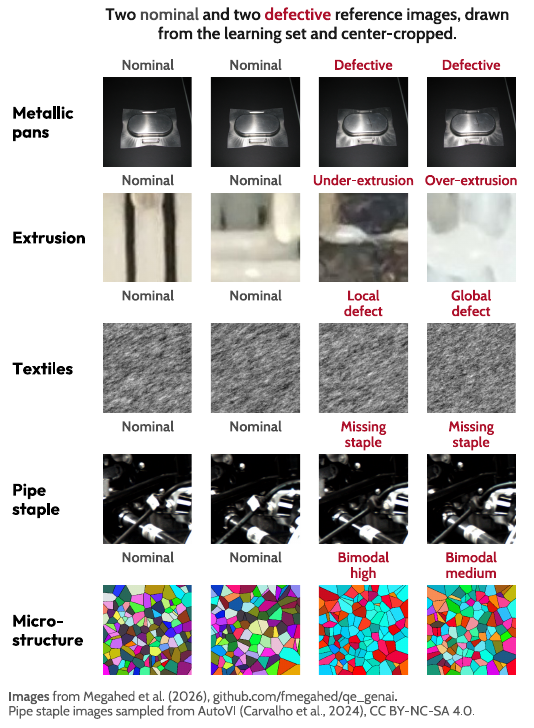}

    \vspace{-0.5\baselineskip}
    
    \caption{An example of the two nominal and two defective reference images per manufacturing case study, used by the decision models and reference-matched CLIP.}
    \label{fig:cmp-examples}

    \vspace{-0.4\baselineskip}
\end{figure}

We evaluate the proposed framework using five manufacturing case studies previously examined in \cite{megahed2026clip}: (a) metallic pan surface nonconformities; (b) over- and under-extrusion in 3D printing; (c) local and global disturbances in stochastic textile surfaces; (d) missing pipe staples in automotive assemblies; and (e) simulated microstructure anomalies. The pipe-staple images were sampled by \cite{megahed2026clip} from the AutoVI dataset \citep{carvalho2024autovi}. Note that the microstructure case study includes three anomaly types (band, bimodal, and single-crystal), each at three severity levels. FIGURE~\ref{fig:cmp-examples} shows the two nominal and two defective reference images used for each case study.

TABLE~\ref{tab:cmp-samples} summarizes the image allocation across the five case studies. For four applications, we use 50 training images per class from \cite{megahed2026clip}. For each application, we select two reference images per class ($R_2$) and use the remaining 48 images per class as the development set ($D$) for agentic description refinement. We also use a separate calibration set ($C$) containing 25 images per class for extrusion, textiles, and microstructure. For pipe staples, the calibration set contains 17 images per class since the defective images were limited. The held-out test sets contain 1,028 images across the five case studies. Although these test images were used in the earlier study of \cite{megahed2026clip}, the agents had no access to them or their results during description refinement.

\begin{table}[htb!]

\vspace{-0.5\baselineskip}

\caption{Image allocation across the five case studies. Counts are per class, except for the test set, which includes both classes.}

\vspace{-0.75\baselineskip}

\label{tab:cmp-samples}
\centering
\small
\begin{tabular}{l@{\hspace{1em}}M@{\hspace{0.8em}}MM@{\hspace{0.8em}}MM}
\toprule
& \multicolumn{1}{c}{Learning Set}
& \multicolumn{2}{c}{Partition of $L$}
& \multicolumn{2}{c}{Evaluation} \\
\cmidrule(lr){2-2}
\cmidrule(lr){3-4}
\cmidrule(lr){5-6}
Case Study & $L$ & $R_2$ & $D$ & $C$ & Test \\
\midrule
Pans           & 10 & 2 & -- & -- & 78 \\
Extrusion      & 50 & 2 & 48 & 25 & 200 \\
Textiles       & 50 & 2 & 48 & 25 & 200 \\
Pipe staples   & 50 & 2 & 48 & 17 & 100 \\
Microstructure & 50 & 2 & 48 & 25 & 450 \\
\bottomrule
\end{tabular}

\vspace{-0.25\baselineskip}

\end{table}

In our experiment, we randomly selected the nominal reference images. When applicable, we also chose defective references to capture different defect types. The reference sets for $k=1,2,4$ are nested and shared across models to ensure consistent comparisons. The pan case study is an exception because only 10 non-test images per class are available. Consequently, we did not perform agentic description refinement and instead used the generic class descriptions. For each configuration with $k$ references per class, we used the remaining $10-k$ images for calibration. Reference-matched CLIP followed the same allocation, while baseline CLIP used leave-one-out scores on $L$, and zero-shot CLIP used all of $L$ for calibration. Thus, unlike the other case studies, the pan application has no independent calibration set.

\subsection{Models and Experimental Controls}
\label{sec:modelsandcontrols}

For our experiments, we used the API-based TypeSafe \texttt{Jev} (\texttt{1.13.0}) and the open-weight Cloudflare \texttt{Clef-Flash} model, which we ran locally using \texttt{SGLang} on an \textit{NVIDIA RTX 5000 Ada Generation} GPU. We also ran \texttt{CLIP ViT-L/14} \citep{radford2021learning} locally on the same GPU for comparison. For the two decision models, we preprocessed the input images by center-cropping them to a square, limiting their dimensions to 768 pixels per side without upscaling, and padding smaller images to at least $280\times280$ pixels before encoding them as JPEG. CLIP used its standard preprocessing procedure, which included resizing and center-cropping to $224\times224$ pixels.

We compared the decision models against the reference-based CLIP approach in \cite{megahed2026clip}, which we refer to as the \textit{CLIP baseline}. In this approach, CLIP computed the maximum cosine similarity between the inspection image and the reference images for each class, then applied softmax to estimate the defect probability. To ensure comparability, we used the same total number of labeled images across the three models: 75 per class for extrusion, textiles, and microstructure; 67 for pipe staples; and 10 for pans. However, these images served different purposes. Specifically, the CLIP baseline used the entire learning set ($L$) as its reference library, whereas the decision models used two reference images per class ($R_2$) and the remaining images to refine their class descriptions. For a given application, all three models applied Platt scaling using the same calibration image set.

In addition to the CLIP baseline, we considered three alternative configurations: (a) \textit{reference-matched CLIP}, which used the same $k$ reference images per class as the decision models; (b) \textit{zero-shot CLIP}, which used the original text descriptions without reference images; and (c) \textit{zero-shot CLIP with refined descriptions}, which replaced the original text descriptions with those selected by \texttt{Jev}. For the two zero-shot configurations, we multiplied the cosine similarity scores by 100 before applying softmax. For conciseness, we report only the results for option (b), which performed better than option (c). Hereafter, we refer to this configuration as \textit{zero-shot CLIP}. We also retained the original, uncalibrated CLIP baseline with a classification threshold of 0.5 for comparison. Note that all alternative configurations used Platt scaling with the calibration set ($C$), including zero-shot CLIP, which did not require reference images at inference. Reference-matched CLIP required fewer labeled images than the primary baseline because it did not involve description refinement.

For the four applications involving agentic description refinement, we conducted eight independent searches, one for each model--application combination. Each search was performed using a separate Claude Code agent (\texttt{claude-opus-5-5, Xhigh} effort), which was provided with the labeled development images, four reference images, the initial class descriptions, and the scoring program. To prevent information leakage, we restricted each agent to its development images, reference images, and scoring program. We also prevented the agents from modifying the scoring program or experimental instructions. 

For each decision model, we evaluated the selected descriptions using $k=0,1,2,4$ reference images per class and the generic descriptions using $k=0,2$. We used $k=2$ for the selected descriptions and did not modify them when we evaluated different numbers of reference images. For the pan application, we used the generic descriptions for all four reference configurations because we did not perform agentic refinement. In total, we evaluated 28 configurations per decision model, each using a separate calibration function and a single evaluation on the held-out test set. We also used file-hash checks to ensure that the descriptions remained unchanged between calibration and testing.

In addition to classification performance, we evaluated the time required to inspect each image. Specifically, we recorded the wall-clock time for \texttt{Jev} and \texttt{Clef-Flash}, including image preparation and model inference. For \texttt{Jev}, the recorded times captured both inference and networking. For CLIP, we measured the time required to (a) load and preprocess each image and extract its features using \texttt{ViT-L/14}, and (b) compare these features with those of the reference images or text descriptions to obtain the classification scores. We timed CLIP after a GPU warm-up run and excluded the one-time encoding of reference images, but included text-prompt encoding for zero-shot classification.  

\subsection{Questions and Evaluation Measures}

To evaluate the proposed framework, we examined four questions: (P1) How does the classification performance of the two decision models using two reference images per class ($k=2$) compare to the CLIP baseline? (P2) Does the agentic refinement of class descriptions improve classification performance? (P3) To what extent do the reference images improve classification performance compared to zero-shot classification? (P4) How does probability calibration affect the reliability of defect probability estimates? For P2, we compared the generic and agentically refined descriptions using $k=2$ reference images per class. For P3, we compared $k=0$ and $k=2$ using the same refined descriptions. We excluded the pan example from P2 because we did not perform agentic refinement, while P3 used the generic descriptions. For P4, we compared the Brier scores and ECEs before and after Platt scaling on the same held-out test images. We also examined isotonic regression to assess how sensitive the calibration results are to the choice of method. For P1–P3, we used calibrated defect probabilities, with a separate calibration function estimated for each experimental configuration.

We evaluated the classification performance using accuracy, sensitivity, specificity, and AUC. To assess the reliability of the estimated defect probabilities, we used the Brier score and expected calibration error (ECE) \cite{guo2017calibration}, calculated using ten equally spaced probability intervals. In our experiments, accuracy and balanced accuracy are equivalent since the test sets contain equal numbers of nominal and defective images. To assess the statistical significance of the observed differences, we used a paired bootstrap approach with 2,000 class-stratified resamples to estimate 95\% confidence intervals for the differences in accuracy and AUC. We also performed exact McNemar tests to compare classification accuracy between methods using the same test images, with Holm adjustments \cite{japkowicz2011evaluating} to account for multiple comparisons. 

\section{Results}

\subsection{Main Classification Performance Results}

Table~\ref{tab:main} summarizes the performance of \texttt{Jev}, \texttt{Clef-Flash}, and three CLIP configurations across the five applications. Using two reference images per class, \texttt{Jev} achieved accuracies ranging from 0.750 to 1.000, while \texttt{Clef-Flash} ranged from 0.710 to 0.990. Both models outperformed reference-matched CLIP across all five applications, although their performance relative to the CLIP baseline using the full reference set varied.

\begin{table*}[htb!]

\caption{Classification performance, inference time, and computational requirements across the five manufacturing case studies. Results are reported after Platt calibration. Bold values indicate the best performance for each case study.}

\vspace{-.75\baselineskip}

\label{tab:main}
\centering
\small
\setlength{\tabcolsep}{3.2pt}

\begin{tabular}{@{}l rrrrrr @{\hspace{14pt}} cr @{\hspace{14pt}} r @{\hspace{14pt}} r@{}}
\toprule
& \multicolumn{6}{c}{Predictive Performance}
& \multicolumn{2}{c}{Inference Time (ms)}
& \multicolumn{1}{c}{API Costs}
& \multicolumn{1}{c}{GPU Memory} \\
\cmidrule(r{6pt}){2-7}
\cmidrule(lr{6pt}){8-9}
\cmidrule(lr{6pt}){10-10}
\cmidrule(l){11-11}

Method & Acc. & Sens. & Spec. & AUC & ECE & Brier
& \lataxis & Median
& \makecell[r]{USD per\\1,000 images}
& \makecell[r]{GB} \\
\midrule

\multicolumn{11}{@{}l}{\textbf{Metallic Pans}}\\
\hspace{1em}\texttt{Jev}, $k=2$
& \textbf{1.000} & \textbf{1.000} & \textbf{1.000} & \textbf{1.000}
& \textbf{0.205} & \textbf{0.056}
& \latglyph{925.1}{944.4}{969.7}{999.5}{1143.2}
& 970 & 0.346 & API \\

\hspace{1em}\texttt{Clef-Flash}, $k=2$
& 0.885 & 0.769 & \textbf{1.000} & \textbf{1.000}
& 0.249 & 0.096
& \latglyph{1159.7}{1183.9}{1211.9}{1243.8}{1301.7}
& 1,212 & 0 & 18.8 \\

\hspace{1em}CLIP baseline
& 0.641 & 0.282 & \textbf{1.000} & 0.999
& 0.290 & 0.183
& \latglyph{63.2}{65.5}{97.9}{110.9}{136.7}
& 98 & 0 & 0.9 \\

\hspace{1em}Reference-matched CLIP, $k=2$
& 0.744 & 0.487 & \textbf{1.000} & 0.988
& 0.255 & 0.146
& \latglyph{99.7}{107.1}{113.3}{128.1}{143.3}
& 113 & 0 & 0.9 \\

\hspace{1em}Zero-shot CLIP
& 0.615 & 0.385 & 0.846 & 0.565
& 0.240 & 0.270
& \latglyph{114.2}{118.2}{124.1}{129.3}{146.4}
& 124 & 0 & 0.9 \\

\addlinespace[3pt]
\multicolumn{11}{@{}l}{\textbf{Extrusion}}\\
\hspace{1em}\texttt{Jev}, $k=2$
& \textbf{0.750} & 0.600 & \textbf{0.900} & \textbf{0.826}
& 0.134 & 0.193
& \latglyph{152.0}{165.6}{175.1}{195.2}{271.7}
& 175 & 0.062 & API \\

\hspace{1em}\texttt{Clef-Flash}, $k=2$
& 0.710 & 0.580 & 0.840 & 0.799
& 0.091 & \textbf{0.191}
& \latglyph{144.3}{167.2}{173.2}{179.9}{188.1}
& 173 & 0 & 18.8 \\

\hspace{1em}CLIP baseline
& 0.715 & \textbf{0.820} & 0.610 & 0.775
& 0.180 & 0.232
& \latglyph{9.2}{9.7}{10.5}{13.8}{19.0}
& 10 & 0 & 0.9 \\

\hspace{1em}Reference-matched CLIP, $k=2$
& 0.670 & 0.600 & 0.740 & 0.743
& 0.090 & 0.216
& \latglyph{9.0}{10.6}{13.3}{14.7}{15.7}
& 13 & 0 & 0.9 \\

\hspace{1em}Zero-shot CLIP
& 0.655 & 0.610 & 0.700 & 0.706
& \textbf{0.069} & 0.223
& \latglyph{14.0}{14.8}{15.9}{19.6}{24.2}
& 16 & 0 & 0.9 \\

\addlinespace[3pt]
\multicolumn{11}{@{}l}{\textbf{Textiles}}\\
\hspace{1em}\texttt{Jev}, $k=2$
& 0.965 & 0.930 & \textbf{1.000} & 0.995
& 0.080 & 0.034
& \latglyph{261.9}{276.0}{289.5}{323.4}{456.3}
& 290 & 0.062 & API \\

\hspace{1em}\texttt{Clef-Flash}, $k=2$
& 0.840 & 0.810 & 0.870 & 0.911
& 0.064 & 0.116
& \latglyph{163.0}{186.4}{193.7}{200.7}{229.8}
& 194 & 0 & 18.8 \\

\hspace{1em}CLIP baseline
& \textbf{0.975} & \textbf{0.970} & 0.980 & \textbf{0.998}
& 0.080 & \textbf{0.029}
& \latglyph{9.4}{9.8}{10.7}{13.0}{16.1}
& 11 & 0 & 0.9 \\

\hspace{1em}Reference-matched CLIP, $k=2$
& 0.685 & 0.640 & 0.730 & 0.753
& 0.066 & 0.196
& \latglyph{9.3}{9.7}{10.4}{12.9}{16.0}
& 10 & 0 & 0.9 \\

\hspace{1em}Zero-shot CLIP
& 0.645 & 0.650 & 0.640 & 0.718
& \textbf{0.059} & 0.214
& \latglyph{14.5}{15.2}{16.5}{20.8}{26.0}
& 16 & 0 & 0.9 \\

\addlinespace[3pt]
\multicolumn{11}{@{}l}{\textbf{Pipe Staples}}\\
\hspace{1em}\texttt{Jev}, $k=2$
& 0.940 & 0.920 & 0.960 & 0.994
& 0.108 & 0.043
& \latglyph{225.1}{250.3}{269.4}{291.8}{342.2}
& 269 & 0.111 & API \\

\hspace{1em}\texttt{Clef-Flash}, $k=2$
& \textbf{0.990} & \textbf{1.000} & \textbf{0.980} & \textbf{1.000}
& 0.108 & \textbf{0.027}
& \latglyph{205.7}{249.4}{257.0}{265.8}{287.1}
& 257 & 0 & 18.8 \\

\hspace{1em}CLIP baseline
& 0.570 & 0.320 & 0.820 & 0.631
& 0.118 & 0.249
& \latglyph{12.4}{12.9}{13.2}{14.2}{16.7}
& 13 & 0 & 0.9 \\

\hspace{1em}Reference-matched CLIP, $k=2$
& 0.670 & 0.680 & 0.660 & 0.686
& \textbf{0.085} & 0.227
& \latglyph{12.3}{12.7}{13.2}{14.1}{18.8}
& 13 & 0 & 0.9 \\

\hspace{1em}Zero-shot CLIP
& 0.640 & 0.580 & 0.700 & 0.709
& 0.087 & 0.228
& \latglyph{17.2}{17.8}{18.2}{19.0}{21.1}
& 18 & 0 & 0.9 \\

\addlinespace[3pt]
\multicolumn{11}{@{}l}{\textbf{Microstructure}}\\
\hspace{1em}\texttt{Jev}, $k=2$
& 0.902 & 0.822 & \textbf{0.982} & 0.959
& 0.059 & 0.075
& \latglyph{225.0}{242.3}{259.6}{288.9}{404.1}
& 260 & 0.114 & API \\

\hspace{1em}\texttt{Clef-Flash}, $k=2$
& 0.867 & 0.751 & \textbf{0.982} & 0.924
& 0.078 & 0.106
& \latglyph{237.1}{247.1}{257.3}{272.4}{306.9}
& 257 & 0 & 18.8 \\

\hspace{1em}CLIP baseline
& \textbf{0.933} & \textbf{0.889} & 0.978 & \textbf{0.990}
& 0.119 & \textbf{0.061}
& \latglyph{10.0}{10.9}{13.2}{15.2}{18.7}
& 13 & 0 & 0.9 \\

\hspace{1em}Reference-matched CLIP, $k=2$
& 0.842 & 0.742 & 0.942 & 0.939
& \textbf{0.057} & 0.106
& \latglyph{9.9}{10.1}{10.3}{10.7}{11.9}
& 10 & 0 & 0.9 \\

\hspace{1em}Zero-shot CLIP
& 0.640 & 0.667 & 0.613 & 0.688
& 0.101 & 0.235
& \latglyph{15.0}{15.5}{15.9}{16.6}{20.5}
& 16 & 0 & 0.9 \\

\bottomrule
\end{tabular}
\end{table*}

The results are encouraging for metallic pans and pipe staples. For pans, \texttt{Jev} correctly classified all 78 test images, while \texttt{Clef-Flash} achieved an accuracy of 0.885. Interestingly, while the CLIP baseline achieved an accuracy of only 0.641, its AUC was 0.999. This suggests that CLIP could distinguish between nominal and defective images, but calibration and the resulting decision threshold affected its classification accuracy. For pipe staples, \texttt{Jev} and \texttt{Clef-Flash} achieved accuracies of 0.940 and 0.990, respectively, compared with 0.570 for CLIP. More importantly, their sensitivities were 0.920 and 1.000, compared with 0.320 for CLIP, indicating substantially fewer missed defects.

For the remaining applications, the differences were generally smaller. \texttt{Jev} had the highest accuracy on extrusion (0.750), while CLIP performed slightly better on textiles (0.975) and microstructure (0.933). Note that the models also differed in their sensitivity and specificity. For example, on extrusion, \texttt{Jev} had higher specificity than CLIP (0.900 versus 0.610), but lower sensitivity (0.600 versus 0.820). Thus, while the decision models performed well across the five applications, their relative performance depended on the application and the considered metrics.

\subsection{Comparison with CLIP}
\label{sec:results-p1}

\begin{figure*}[htb!]
    \centering
    \includegraphics[width=.94\linewidth, frame]{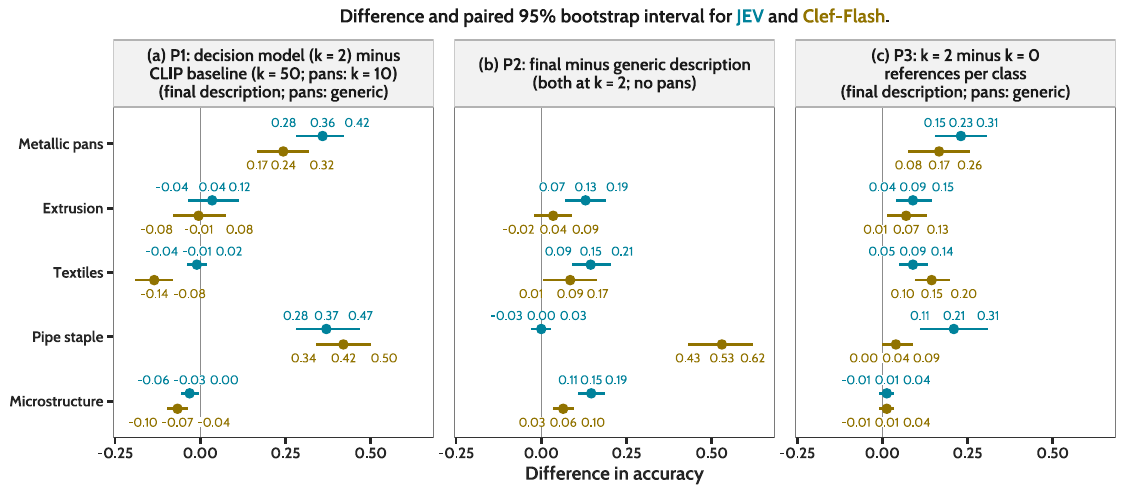}

    \vspace{-0.5\baselineskip}
    
    \caption{Paired accuracy differences (95\% bootstrap confidence intervals) for \texttt{Jev} and \texttt{Clef-Flash}: (a) versus CLIP, (b) refined versus generic descriptions, and (c) two versus zero reference images per class.}
    \label{fig:paired}
\end{figure*}

FIGURE~\ref{fig:paired}(a) compares the classification accuracy of \texttt{Jev} and \texttt{Clef-Flash} against the CLIP baseline from our previous study, which used 50 reference images per class (10 for metallic pans). Overall, the results for \texttt{Jev} are particularly encouraging. Using only two reference images per class at inference, \texttt{Jev} achieved comparable or better classification accuracy than CLIP across the five applications. We observed statistically significant improvements for metallic pans and pipe staples, while the Holm-adjusted McNemar tests did not identify significant differences for the remaining applications. \texttt{Clef-Flash} also outperformed CLIP on metallic pans and pipe staples, but achieved lower accuracy on textiles and microstructure.

The results are less favorable on textiles and microstructure. \texttt{Jev} performs similarly to CLIP on textiles, with an accuracy difference of $-0.010$ (95\% bootstrap CI: $-0.040$-$0.020$). On microstructure, its difference is $-0.031$ (95\% bootstrap CI: $-0.058$-$-0.004$), while the Holm-adjusted McNemar test is not significant ($p=0.168$). \texttt{Clef-Flash} performs significantly worse than CLIP on both applications, with accuracy differences of $-0.135$ and $-0.067$, respectively (adjusted $p<0.001$ for both). 

For extrusion, neither decision model shows a statistically significant accuracy improvement over CLIP. However, as discussed earlier, similar classification accuracies do not necessarily imply similar sensitivity and specificity. This distinction is particularly important in manufacturing inspection, where the consequences of missing a defective product may be different from those of rejecting a conforming product.

\subsection{Description Refinement and Reference Images}
\label{sec:results-refinement}

FIGURE~\ref{fig:paired}(b) summarizes the effects of using agentic AI to refine the semantic descriptions. Overall, we observed statistically significant improvements in classification accuracy for six of the eight model–application combinations, as indicated by the paired 95\% bootstrap confidence intervals. The largest improvement was for \texttt{Clef-Flash} on pipe staples, where accuracy increased by 0.53 (95\% CI: 0.43--0.62). Except for \texttt{Jev} on pipe staples and \texttt{Clef-Flash} on extrusion, moving from generic descriptions to more precise, application-specific descriptions yielded statistically significant accuracy improvements.

To illustrate the refinement process, TABLE~\ref{tab:description-long} provides an example for \texttt{Jev} on extrusion. We started with generic descriptions that distinguished between parts with and without visible manufacturing defects, yielding a development AUC of 0.59. In the first iteration, the agent rewrote both descriptions to include extrusion-specific characteristics, such as material width, continuity, gaps, and excess plastic. These changes to both the nominal and defective descriptions increased the AUC to 0.74. With this increase in the AUC trajectory, the AI agent made only small changes to specific words in the third iteration (as evidenced by our bolded words for new additions and strikethrough words for removal), which further improved the AUC to 0.78. In our opinion, this dichotomy (making substantial changes when classification performance is subpar and targeted changes as the AI agent accumulates more information) is an interesting result, as it mimics how some experts would refine their descriptions.

\begin{table}[htb!]
\caption{\label{tab:description-long}The descriptions that the Jev description loop kept for extrusion, with Git-style track changes.}

\vspace{-0.75\baselineskip}

\centering
\small 
\begin{threeparttable}
\begin{tabularx}{\columnwidth}{@{}l>{\raggedright\arraybackslash}X@{}}
\toprule
\multicolumn{2}{@{}l}{\textbf{Candidate 1}, development AUC 0.59} \\
Nominal & No visible manufacturing defects; the part appears conforming. \\
Defective & At least one visible manufacturing defect or anomaly. \\
\addlinespace
\multicolumn{2}{@{}l}{\textbf{Candidate 2}, development AUC 0.74} \\
Nominal & \textbf{A close-up of a 3D printing nozzle and freshly extruded plastic; the extruded material forms a smooth, even layer of steady width that lies flat on the print surface.} \\
Defective & \textbf{A close-up of a 3D printing nozzle and extruded plastic where the amount of material is wrong: either a thin, stringy or broken strand with dark gaps showing through, or a bulging, blobby, overly wide mass of plastic.} \\
\addlinespace
\multicolumn{2}{@{}l}{\textbf{Candidate 3}, development AUC 0.78} \\
Nominal & A close-up of a 3D printing nozzle and freshly extruded plastic; the extruded material forms a smooth, even layer \textcolor{black}{\textbf{or strand}} of steady width \textcolor{black!60}{\sout{that lies flat on the print surface.}} \textcolor{black}{\textbf{with straight, clean edges. The background around it may be dark or light.}} \\
Defective & A close-up of a 3D printing nozzle and extruded plastic where the amount of material is wrong: either a thin, stringy \textcolor{black!60}{\sout{or broken strand with dark gaps showing through,}} \textcolor{black}{\textbf{strand with breaks or gaps within the plastic itself,}} or a bulging, blobby, overly wide mass of plastic. \\
\bottomrule
\end{tabularx}
\end{threeparttable}

\vspace{-.75\baselineskip}

\end{table}

FIGURE~\ref{fig:paired}(c) compares zero-shot classification ($K=0$) with using two reference images per class ($K=2$). For \texttt{Jev}, adding reference images improved accuracy in four of the five applications, with paired 95\% bootstrap confidence intervals excluding zero; microstructure was the only exception. For \texttt{Clef-Flash}, the confidence intervals excluded zero in three applications, while pipe staples and microstructure showed no clear improvement. These results are consistent with what we would expect from the genAI literature, where improving the context helps models make more informed decisions.

\subsection{Probability Calibration and Computational Time}
\label{sec:results-calibration}

\begin{figure}[htb!]    
    \centering
    \includegraphics[width=0.99\linewidth, frame, trim={0 0 0 .55in}, clip]{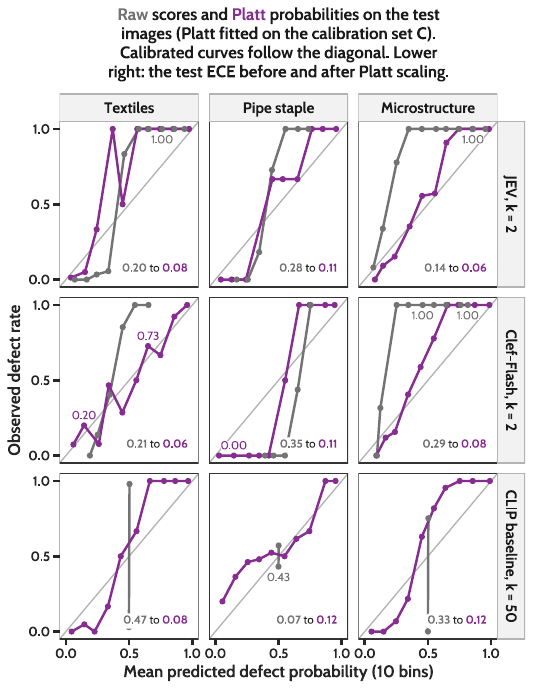}

    \vspace{-0.5\baselineskip}
    
    \caption{Reliability diagrams comparing raw (gray) and Platt-calibrated (purple) defect probabilities for \texttt{Jev}, \texttt{Clef-Flash}, and CLIP across three sample case studies (see supplementary materials for full results). Values indicate test ECE before and after calibration.}
    \label{fig:reliability}

    \vspace{-.5\baselineskip}
\end{figure}

Figure~\ref{fig:reliability} shows the reliability diagrams before and after Platt calibration for three representative applications. Overall, calibration improved the quality of the estimated defect probabilities, reducing the Brier score in 13 of the 15 model–application combinations. For ECE, we observed improvements for \texttt{Jev} and \texttt{Clef-Flash} across all six configurations shown in the figure. For CLIP, ECE decreased for textiles and microstructure but increased slightly for pipe staples.

The metallic pan application illustrates why we need to distinguish between classification accuracy, AUC, and calibration. While the CLIP baseline achieved an AUC of 0.999, its classification accuracy decreased from 0.910 using the raw 0.5 threshold to 0.641 after calibration. This application lacked an independent calibration set. As such, the results illustrate how calibration can be sensitive to the amount of data available, even when a model can distinguish between nominal and defective images.

We also considered isotonic regression as an alternative to Platt scaling. The results, provided in Table 14 under the ``Analysis notebook'' tab in the Replication Materials, show that isotonic regression achieved lower ECE in eight of the ten decision-model configurations with two reference images per class. However, we also observed substantially higher log loss in some configurations. Thus, while isotonic regression may provide better ECE, it does not necessarily improve all measures of probability reliability.

Finally, Table~\ref{tab:main} summarizes the inference times and computational requirements. Excluding metallic pans, the median inference times ranged from 175 to 290 ms for \texttt{Jev} and 173 to 257 ms for \texttt{Clef-Flash}, compared to approximately 10--13 ms for CLIP. All three models required more time for metallic pans. We accessed \texttt{Jev} through a remote API, with estimated charges ranging from \$0.062 to \$0.346 per 1,000 images. In contrast, we ran \texttt{Clef-Flash} and CLIP locally on a GPU, requiring about 18.8 GB and 0.9 GB of memory, respectively. The costs reported for \texttt{Clef-Flash} and CLIP exclude the initial investment in GPU hardware and the increased electricity consumption from the resulting computations.

\section{Discussion and Concluding Remarks}

\subsection{Summary of Main Contributions}

Our findings suggest that \textit{System One} models offer a promising approach to 100\% product inspection, particularly for low-volume and high-yield applications where collecting large amounts of labeled data may not be practical. Across five case studies, we found that \texttt{Jev} and \texttt{Clef-Flash} can achieve competitive classification performance with as few as $k=2$ reference images per class during inference. Our use of a separate development set to refine semantic descriptions allowed us to assess a relatively rare, practical worst-case scenario when practitioners cannot provide precise descriptions of the classes they inspect. From our analyses, we identified two factors that can improve classification performance: (a) more informative, application-specific class descriptions, and (b) increasing the number of reference images, especially when compared to zero-shot \textit{System One} implementations. 

Another important finding relates to calibrating the models' output probabilities. We show that Platt scaling reduced the Brier score in 13 of the 15 model-application combinations. This is an important result for applications when the reliability of the reported probabilities is used in decision-making, e.g., if low-confidence decisions should warrant additional inspection. 

To facilitate the replication of our experiments and the evaluation of \textit{System One} models in other applications, we developed the ``decisions'' \faRProject \, package. We hope this encourages further research into using such models in manufacturing applications.

\subsection{Advice to Practitioners}

Our proposed framework is particularly useful in manufacturing applications where collecting enough images to train traditional computer vision models is impractical. In these settings, practitioners can potentially achieve good classification performance with as few as two reference images per class and a meaningful description of the defects they are trying to identify. This makes our approach relatively easy to implement.

Another point to consider is that we intentionally started with generic text descriptions and used \textasciitilde50 images to improve them through our agentic AI approach. However, in many cases, practitioners may have a much better understanding of their processes and the types of defects they may encounter. As such, they may be able to provide sufficiently informative descriptions from the beginning, without needing our agentic loop approach. To help practitioners evaluate the proposed approach on their own applications, we have made a free, web-based app available where they can experiment with their own images (see \url{https://huggingface.co/spaces/fmegahed/decisions}).

\subsection{Limitations and Future Research Directions}

Our study has several limitations. First, although we considered five manufacturing applications, the images were collected under relatively stable conditions. It is unclear how well the proposed models would perform over longer periods in production environments where lighting, equipment, and other operating conditions may change. Future research should examine the robustness of these models under such conditions. However, given the relatively small number of reference images required, it may be possible to adapt these models to changing conditions without collecting large amounts of new data.

Second, our experiments focused on binary defect classification. We have not examined how well the proposed approach would perform in multi-class settings or whether it can identify where a defect occurs within an image. For multi-class problems, one potential challenge is the number of reference images that can be passed to the TypeSafe/\texttt{Jev} API or stored in context for other \textit{System One} models. This may require dividing the classes into smaller groups or developing a hierarchical classification approach. Combining the ``Choice'' or ``Noul'' primitives with the recently introduced ``Box'' primitive \citep{typesafe2026primitives,typesafe2026box} could extend our framework to defect localization. Finally, these models remain largely black boxes. While we have shown that changes to semantic text descriptions can improve classification performance, we do not fully understand why some descriptions work better than others. Our agentic AI approach provides one way to automate this process, but it does not explain the underlying model behavior. We also plan to investigate how to incorporate conformal prediction into our framework to better characterize the uncertainty associated with the resulting classifications \citep{angelopoulos2023conformal}.

\subsection{Concluding Remarks}

In this study, we show how \textit{System One} decision models, such as \texttt{Jev} and \texttt{Clef-Flash}, can enable 100\% inspection in low-volume, customized, and/or high-yield manufacturing applications by combining semantic text descriptions with as few as $k=2$ reference images. We apply our proposed framework to five manufacturing case studies and compare its performance against three CLIP-based baselines. Our results show that \texttt{Jev} (and often \texttt{Clef-flash}) can achieve comparable or better classification accuracies across these applications. We also demonstrate that the text/class descriptions can be improved automatically with an agentic AI loop, and that increasing the number of reference images can further improve classification performance. To support research and practical adoption, we provide the \faRProject \, ``decisions'' package and an accompanying no-code web app.

\section*{Acknowledgments}
The authors acknowledge the \textit{Farmer School of Business} and the \textit{Raymond E. Glos Professorship in Business} for providing the computational resources and monetary support for this research.

\section*{Disclosure of AI Use}
We used Claude Code (Anthropic; versions 2.1.304–2.1.312, using Claude Opus 5.5, Sonnet 5.5, and Haiku 5.5) through a command-line interface on a Windows 11 computer between October 20 and 28, 2026. Under the supervision of Dr. Megahed, Claude Code generated the project's computational code, including the \faRProject \, scripts used for FIGURES~3–5, and translated our \faRProject \, results into \LaTeX \, tables (TABLES~1–3). Claude Opus 5.5 was also used in the automated description-generation experiments, as described in Sections \ref{sec:cmp-agent} and \ref{sec:modelsandcontrols}.

Additional AI tools included OpenAI Codex (GPT-6 Astra, xhigh reasoning effort, v 0.163.0) for refining the \LaTeX \, code used to generate FIGURE~2 on October 24–25. Finally, ChatGPT.com (OpenAI; Sol 5.6, medium effort) and the Grammarly Chrome Extension (Superhuman Platform Inc., v. 14.1342.0) were both used for language editing and readability improvements from October 20 to 29, 2026.

The authors were responsible for the research design, methodological decisions, code review/tests, interpretation of results, and final manuscript content. Dr. Megahed created a Claude ``replication-materials'' skill, which documented the \textasciitilde 300 decisions made when interacting with Claude Code. See \url{https://github.com/fmegahed/claude-skills} for the Claude Skill and the next subsection for the detailed record of AI involvement.

\section*{Replication Materials}
While GenAI has the potential to accelerate scientific research, there are growing concerns about its broader implications for scientific discovery and research integrity \citep[e.g., see][]{hao2026artificial,hochberg2026ai}. We believe AI-augmented approaches can be impactful in manufacturing applications; however, they must be implemented transparently so that the research can be replicated. Therefore, our supplementary website (\url{https://fmegahed.github.io/decision-models-for-industrial-image-qc/}, generated by our Claude skill) serves as an AI-augmented laboratory notebook, documenting (a) the \textasciitilde 300 research decisions made throughout this study, including who proposed and approved each decision, when it was made, and the extent of AI involvement; (b) GitHub code repository, with a full commit history; and (c) our Quarto (`.qmd') notebook, capturing our code, results, and artifacts included in this paper (as well as additional supplementary figures that we could not include due to space considerations).

\bibliographystyle{asmeconf}
\bibliography{refs}

\begin{thebibliography}{10}
\newcommand{\enquote}[1]{``#1''}
\providecommand{\url}[1]{\texttt{#1}}
\providecommand{\urlprefix}{URL }
\expandafter\ifx\csname urlstyle\endcsname\relax
  \providecommand{\doi}[1]{DOI \discretionary{}{}{}#1}\else
  \providecommand{\doi}{DOI \discretionary{}{}{}\begingroup \urlstyle{rm}\Url}\fi
\providecommand{\eprint}[2][]{\urlprefix\url{#1#2}}

\bibitem{colosimo2024statistical}
Colosimo, Bianca~M, Jones-Farmer, L~Allison, Megahed, Fadel~M, Paynabar, Kamran, Ranjan, Chitta and Woodall, William~H.
\newblock \enquote{Statistical process monitoring from industry 2.0 to industry 4.0: Insights into research and practice.}
\newblock \emph{Technometrics} Vol.~66 No.~4 (2024): pp. 507--530.
\newblock \doi{10.1080/00401706.2024.2327341}.

\bibitem{megahed2011review}
Megahed, Fadel~M., Woodall, William~H. and Camelio, Jaime~A.
\newblock \enquote{A Review and Perspective on Control Charting with Image Data.}
\newblock \emph{Journal of Quality Technology} Vol.~43 No.~2 (2011): pp. 83--98.
\newblock \doi{10.1080/00224065.2011.11917848}.

\bibitem{jeong2023winclip}
Jeong, Jongheon, Zou, Yang, Kim, Taewan, Zhang, Dongqing, Ravichandran, Avinash and Dabeer, Onkar.
\newblock \enquote{{WinCLIP}: Zero-/Few-Shot Anomaly Classification and Segmentation.}
\newblock \emph{2023 IEEE/CVF Conference on Computer Vision and Pattern Recognition (CVPR)}: pp. 19606--19616. 2023.
\newblock \doi{10.1109/CVPR52729.2023.01878}.

\bibitem{psarommatis2026zero}
Psarommatis, Foivos, Azamfirei, Victor, Solina, Vittorio, May, Gokan and Jun, Hongbae.
\newblock \enquote{Zero Defect Manufacturing – from vision to integrated industrial reality.}
\newblock \emph{International Journal of Production Research} Vol.~64 No.~17 (2026): pp. 7261--7268.
\newblock \doi{10.1080/00207543.2026.2707319}.

\bibitem{hu2013evolving}
Hu, S.~Jack.
\newblock \enquote{Evolving Paradigms of Manufacturing: From Mass Production to Mass Customization and Personalization.}
\newblock \emph{Procedia CIRP} Vol.~7 (2013): pp. 3--8.
\newblock \doi{10.1016/j.procir.2013.05.002}.

\bibitem{radford2021learning}
Radford, Alec, Kim, Jong~Wook, Hallacy, Chris, Ramesh, Aditya et~al.
\newblock \enquote{Learning Transferable Visual Models From Natural Language Supervision.}
\newblock Meila, Marina and Zhang, Tong (eds.). \emph{Proceedings of the 38th International Conference on Machine Learning}, Vol. 139: pp. 8748--8763. 2021. PMLR.
\newblock \urlprefix\url{https://proceedings.mlr.press/v139/radford21a.html}.

\bibitem{megahed2026clip}
Megahed, Fadel~Mounir, Chen, Ying-Ju, Colosimo, Bianca~Maria, Grasso, Marco, Jones-Farmer, L.~Allison, Knoth, Sven, Sun, Hongyue and Zwetsloot, Inez~M.
\newblock \enquote{Adapting {OpenAI}'s {CLIP} Model for Few-Shot Image Inspection in Manufacturing Quality Control: An Expository Case Study With Multiple Application Examples.}
\newblock \emph{Journal of Quality Technology} Vol.~58 No.~4 (2026): pp. 415--441.
\newblock \doi{10.1080/00224065.2026.2640171}.

\bibitem{typesafe2026introducing}
{TypeSafe AI}.
\newblock \enquote{Introducing System One Models \& Jev.}
\newblock \url{https://typesafe.ai/blog/introducing-system-one-models-and-jev} (2026).
\newblock Last accessed on October 25, 2026.

\bibitem{cloudfare2025clef}
{Cloudflare}.
\newblock \enquote{Cloudflare/clef-flash.}
\newblock \url{https://huggingface.co/Cloudflare/clef-flash} (2026).
\newblock Last accessed on October 25, 2026.

\bibitem{typesafe2026primitives}
{TypeSafe AI}.
\newblock \enquote{Primitives (Questions).}
\newblock \url{https://docs.typesafe.ai/primitives} (2026).
\newblock Last accessed on October 26, 2026.

\bibitem{typesafe2026box}
{TypeSafe AI}.
\newblock \enquote{Box - {TypeSafe}.}
\newblock \url{https://docs.typesafe.ai/primitives/box} (2026).
\newblock Last accessed on October 26, 2026.

\bibitem{karpathy2026autoresearch}
Karpathy, Andrej.
\newblock \enquote{Autoresearch.}
\newblock \url{https://github.com/karpathy/autoresearch} (2026).
\newblock Software repository. Accessed October 22, 2026.

\bibitem{platt1999probabilistic}
Platt, John~C.
\newblock \enquote{Probabilistic Outputs for Support Vector Machines and Comparisons to Regularized Likelihood Methods.}
\newblock Smola, Alexander~J., Bartlett, Peter~L., Sch{\"o}lkopf, Bernhard and Schuurmans, Dale (eds.). \emph{Advances in Large Margin Classifiers}.
\newblock MIT Press, Cambridge, MA (1999): pp. 61--74.

\bibitem{zadrozny2002transforming}
Zadrozny, Bianca and Elkan, Charles.
\newblock \enquote{Transforming Classifier Scores Into Accurate Multiclass Probability Estimates.}
\newblock \emph{Proceedings of the Eighth ACM SIGKDD International Conference on Knowledge Discovery and Data Mining}: pp. 694--699. 2002.
\newblock \doi{10.1145/775047.775151}.

\bibitem{carvalho2024autovi}
Carvalho, Philippe, Lafou, Meriem, Durupt, Alexandre, Leblanc, Antoine and Grandvalet, Yves.
\newblock \enquote{The Automotive Visual Inspection Dataset ({AutoVI}): A Genuine Industrial Production Dataset for Unsupervised Anomaly Detection.} (2024).
\newblock \doi{10.5281/zenodo.10459003}.

\bibitem{guo2017calibration}
Guo, Chuan, Pleiss, Geoff, Sun, Yu and Weinberger, Kilian~Q.
\newblock \enquote{On Calibration of Modern Neural Networks.}
\newblock \emph{Proceedings of the 34th International Conference on Machine Learning}, Vol.~70: pp. 1321--1330. 2017.
\newblock \urlprefix\url{https://proceedings.mlr.press/v70/guo17a.html}.

\bibitem{japkowicz2011evaluating}
Japkowicz, N. and Shah, M.
\newblock \emph{Evaluating Learning Algorithms: A Classification Perspective}.
\newblock Cambridge University Press (2011).

\bibitem{angelopoulos2023conformal}
Angelopoulos, Anastasios~N and Bates, Stephen.
\newblock \enquote{Conformal prediction: A gentle introduction.}
\newblock \emph{Foundations and Trends in Machine Learning} Vol.~16 No.~4 (2023): pp. 494--591.
\newblock \doi{10.1561/2200000101}.

\bibitem{hao2026artificial}
Hao, Qianyue, Xu, Fengli, Li, Yong and Evans, James.
\newblock \enquote{Artificial intelligence tools expand scientists’ impact but contract science’s focus.}
\newblock \emph{Nature} Vol. 649 No. 8099 (2026): pp. 1237--1243.
\newblock \doi{10.1038/s41586-025-09922-y}.

\bibitem{hochberg2026ai}
Hochberg, Michael~E and Thrall, Peter~H.
\newblock \enquote{AI will reorganize science. Will research remain a human enterprise?}
\newblock \emph{Proceedings of the National Academy of Sciences} Vol. 123 No.~31 (2026): p. e2610088123.
\newblock \doi{10.1073/pnas.2610088123}.

\end{thebibliography}

\end{document}